\def\year{2022}\relax
\documentclass[letterpaper]{article} 
\usepackage{aaai22}  
\usepackage{times}  
\usepackage{helvet}  
\usepackage{courier}  
\usepackage[hyphens]{url}  
\usepackage{graphicx} 

\usepackage{scalerel} 
\usepackage{enumitem}
\usepackage{bbm}
\usepackage{multirow}
\usepackage{subcaption}
\usepackage{booktabs}
\usepackage{amsmath}
\usepackage{tabularx}
\usepackage{ragged2e}
\usepackage{amsfonts}
\usepackage{xcolor}

\usepackage{natbib}  
\usepackage{caption} 
\DeclareCaptionStyle{ruled}{labelfont=normalfont,labelsep=colon,strut=off} 
\usepackage{algorithm}
\usepackage{algorithmic}

\usepackage{newfloat}
\usepackage{listings}
\floatstyle{ruled}
\newfloat{listing}{tb}{lst}{}
\floatname{listing}{Listing}
\title{Continual Learning for Unsupervised Anomaly Detection \\ in Continuous Auditing of Financial Accounting Data}

\author{
      Hamed Hemati\equalcontrib \textsuperscript{\rm 1}, 
      Marco Schreyer\equalcontrib \textsuperscript{\rm 1},
      Damian Borth \textsuperscript{\rm 1}\\
 }
\affiliations{
        \textsuperscript{\rm 1}University of St. Gallen (HSG), St. Gallen, Switzerland\\

}

\usepackage{bibentry}

\begin{document}

\maketitle

\begin{abstract}
International audit standards require the direct assessment of a financial statement's underlying accounting journal entries. Driven by advances in artificial intelligence, deep-learning inspired audit techniques emerged to examine vast quantities of journal entry data. However, in regular audits, most of the proposed methods are applied to learn from a comparably stationary journal entry population, e.g., of a financial quarter or year. Ignoring situations where audit relevant distribution changes are not evident in the training data or become incrementally available over time. In contrast, in continuous auditing, deep-learning models are continually trained on a stream of recorded journal entries, e.g., of the last hour. Resulting in situations where previous knowledge interferes with new information and will be entirely overwritten. This work proposes a continual anomaly detection framework to overcome both challenges and designed to learn from a stream of journal entry data experiences. The framework is evaluated based on deliberately designed audit scenarios and two real-world datasets. Our experimental results provide initial evidence that such a learning scheme offers the ability to reduce false-positive alerts and false-negative decisions.
\end{abstract}

\section{Introduction}
\label{sec:introduction}

\begin{figure}[t!]
	\hspace*{0.0cm} \includegraphics[width=9.2cm, angle=0, trim={4.5cm 1.0cm 0.0 0.0}]{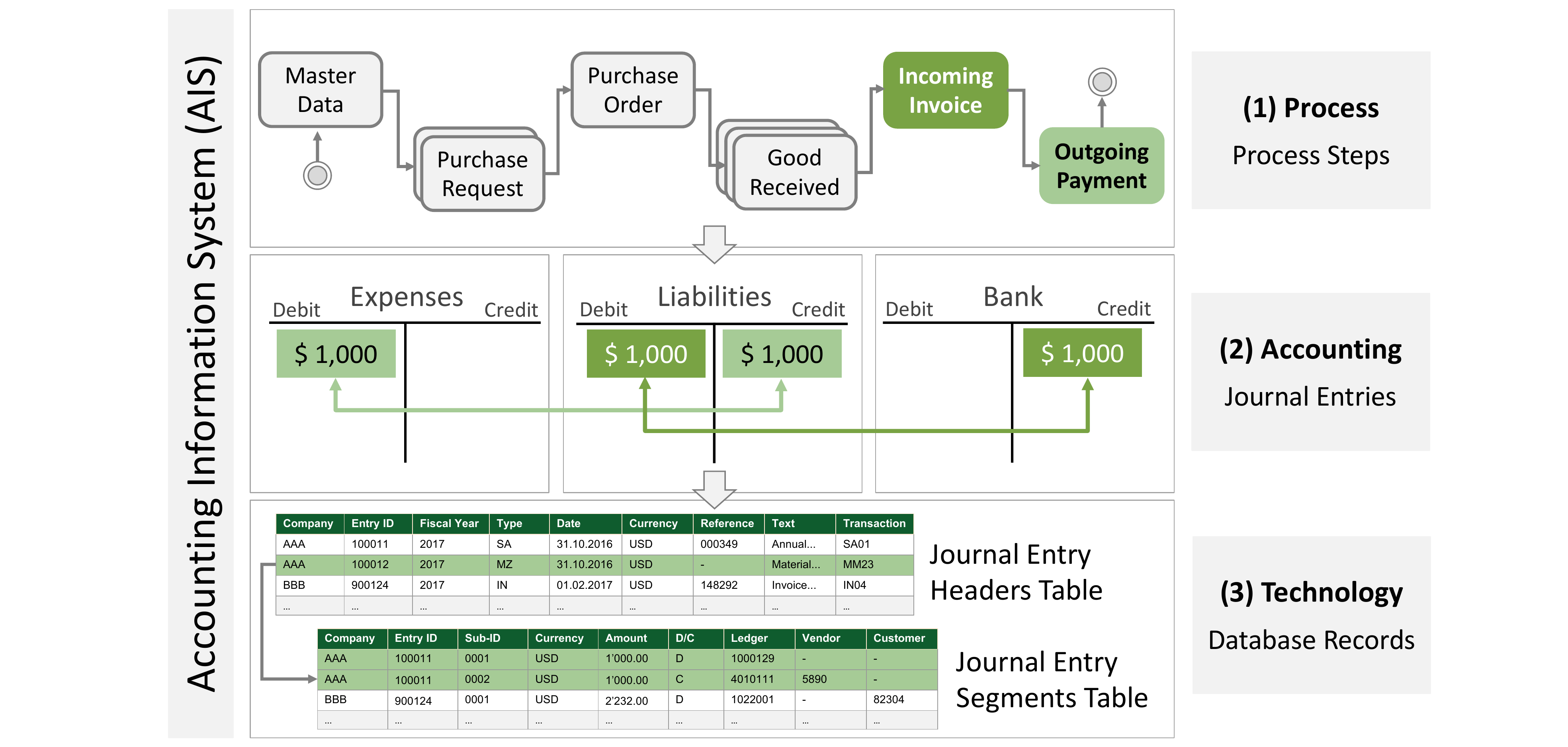}
	\vspace{-3mm}
	\caption{Exemplary view of an Accounting Information System (AIS) that records distinct layer of abstractions, namely (1) the business process, (2) the accounting and (3) technical journal entry information in designated tables.}
	\label{fig:ais_system}
	\vspace{-4mm}
\end{figure}

According to the \textit{International Standards in Auditing (ISA)}, auditors are obliged to collect reasonable assurance that an issued financial statement is free from fraud that causes a material misstatement \citep{sas99, ifac2009}. At the same time, the \textit{Association of Certified Fraud Examiners (ACFE)} estimates that organizations lose 5\% of their annual revenues due to fraud \cite{ACFE2020}. The term \textit{fraud} refers to \textit{`the abuse of one's occupation for personal enrichment through the deliberate misuse of an organization's resources or assets'} \cite{Wells2017}. A similar study revealed that approx. 30\% of the respondents experienced losses between \$100,000 and \$5 million due to fraud \cite{PWC2020}. The study also showed that financial statement fraud caused the highest median loss of the surveyed fraud schemes.\footnote{The ACFE study encompasses an analysis of 2,504 cases of occupational fraud surveyed between July 2019 and September 2019 in 125 countries. The PwC study encompasses over 5,000+ respondents that experienced economic crime in the last 24 months.} To detect such fraud, the ISA demands auditors to examine a statement's underlying accounting transactions, usually referred to as \textit{journal entries} \citep{caq2008, ifac2009}. Nowadays, organizations record vast quantities of journal entries in \textit{Accounting Information Systems (AIS)} or more general \textit{Enterprise Resource Planning (ERP)} systems \citep{grabski2011}. Figure \ref{fig:ais_system} depicts an exemplary hierarchical view of the journal entry recording process in designated database tables of an AIS system. In general, such recorded entries originate from non-stationary activities due to the dynamics of organizational processes, e.g., the (i) introduction, (ii) redesign, or (iii) discontinuation of business processes. Therefore, the detection of fraud in non-stationary journal entry distributions remains a challenging task requiring significant efforts and resources. 

In order to conduct fraud, perpetrators need to deviate from regular journal entry posting patterns. Such a divergence is then recorded in a small fraction of journal entries that exhibit \textit{`anomalous'} attribute values. Nowadays, auditors apply a wide range of \textit{Computer Assisted Audit Techniques (CAATs)} when examining journal entries. These techniques often encompass rule-based analyses, such as the detection of unusual posting times \cite{fay2017} or multiple vendor bank account changes \cite{coderre2009}. Furthermore, statistical analyses are applied, such as the \textit{Benford's Law} \cite{Benford1938} or extreme value analysis \cite{bolton2002}. Driven by the rapid technological advances of artificial intelligence, deep-learning \citep{LeCun2015} enabled audit techniques have recently emerged in financial audits \citep{sun2019, nonnenmacher2021b}.

At the same time, the application of deep-learning in modern audit engagements results, among others, in the challenge of (i) \textit{stationary training setups} and (ii) \textit{catastrophic forgetting}. The first challenge originates from the situation that in \textit{regular auditing} deep-learning models are learned from a limited and therefore \textit{stationary population} of journal entries. Typically the training only encompasses entries posted throughout the in-scope audit period, e.g., the current financial quarter or year. Such a learning setup exhibits the drawback that relevant distribution changes might not be evident in the training data. The second challenge originates from the fact that deep-learning based audit models are prone to \textit{Catastrophic Forgetting} (CF) \cite{mccloskey1989catastrophic}, i.e., training a model with new information interferes with previously learned knowledge. Typically, in \textit{continuous auditing} \cite{vasarhelyi1991, vasarhelyi2004} the model training encompasses a constant stream of entries recorded within incremental time intervals, e.g., the recent hour or day. This phenomenon leads, in the worst-case, to a situation in which previous knowledge will be entirely overwritten by new one. 

In order to mitigate false-positive alerts or false-negative decisions in real-world audits, it is paramount to learn important audit relevant information that becomes incrementally available over time. For example, an organizational change in previous financial years might be relevant when concluding on a recently detected audit finding. This holds in particular for applying unsupervised anomaly detection techniques to audit large-scale accounting data. Recently, the idea of \textit{Continual Learning} (CL), also referred to as \textit{lifelong learning}, regained attention \cite{chen2018lifelong}. CL denotes a learning paradigm in which the assumption is made that data will become available over time. We think accommodating new knowledge over time while retaining previously learned information is essential for future deep-learning enabled audit techniques. In this work, inspired by the successes of CL \cite{sun2019, zhai2019, lee2020}, we investigate whether such a CL setup can be utilized in financial audits to detect accounting anomalies? In summary, we present the following contributions:

\begin{itemize}

\item \textit{Learnability} - We propose and evaluate a novel learning framework to continuously learn from real-world journal entry for unsupervised anomaly detection. 

\vspace{-1mm}

\item \textit{Transferability} - We demonstrate that the continual learning techniques outperform the re-training on an exhaustive dataset and the sequential fine-tuning. 


\end{itemize}




The remainder of this work is structured as follows: First, we provide an overview of related work. Following, we present the proposed continual learning framework for unsupervised anomaly detection in financial audits. Afterwards, the experimental setup is described and results are outlined. The work concludes with a discussion and summary.

\begin{figure*}[ht!]
    \center
    \includegraphics[height=6.5cm,trim=0 0 0 12, clip]{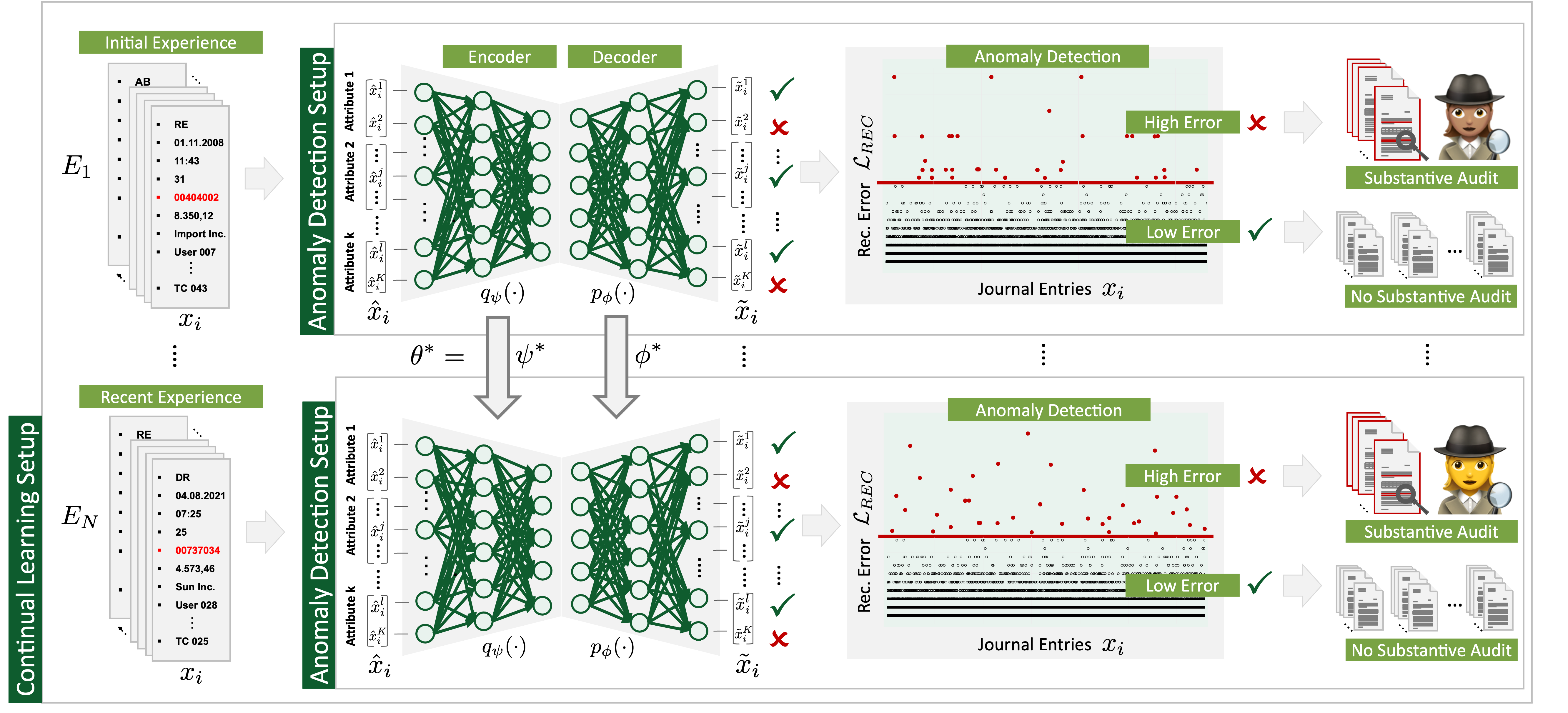}
    \vspace{-2mm}
    \caption{The proposed continual anomaly detection framework that continuously learns from a stream of accounting data experiences $\{E_{i}\}_{i=1}^{N}$ (e.g., financial quarters or years) by accommodating new information while preserving previous knowledge.}
    \label{fig:architecture}
    \vspace{-4mm}
\end{figure*}

\section{Related Work}
\label{sec:related_work}

Lately, techniques that build on (deep) machine learning have been gradually applied to different audit tasks \citep{sun2019, cho2020}. In parallel, the idea of continual learning triggered a sizable body of machine learning research \citep{parisi2019, delange2021}. In the following, we focus our literature review on (i) CAATs applied in financial audits and (ii) continual deep-learning techniques.

\vspace{1mm}

\textbf{Computer Assisted Audit Techniques:} With the evolution of AIS systems, the development of CAATs has been of central relevance in auditing. Various techniques have been proposed to audit structured accounting data, such as (un-) supervised machine-learning approaches \citep{Bay2002, McGlohon2009, Jans2010, thiprungsri2011, Argyrou2012, Argyrou2013, hu2020}, statistical analyzes \citep{Debreceny2010, Seow2016}, rule-based procedures \citep{Khan2009, Khan2010, Khan2014, Islam2010}, business process mining \citep{Jans2011, jans2014, werner2015, werner2016}, and the combination thereof \citep{baader2018, stephan2021}. Recently, deep-learning techniques have emerged in auditing. Such techniques encompass, autoencoder neural networks \citep{schreyer2017, schultz2020, nonnenmacher2021a, zupan2020, schreyer2020} to detect accounting anomalies and learn representative audit samples. Lately, self-supervised learning has been proposed to learn multi-task audit models \citep{schreyer2021}.

\vspace{1mm}

\textbf{Continual Deep-Learning Techniques:} The challenge of CF in training neural networks, has been studied extensively \citep{french1999catastrophic, diaz2018don, toneva2018empirical, lesort2021understanding}. To mitigate CF, several approaches have been proposed, broadly classified as (i) \textit{rehearsal}, (ii) \textit{regularization}, and (iii) \textit{dynamic architecture} techniques. \textit{Rehearsal} based techniques \citep{rolnick2018experience, isele2018selective, chaudhry2019continual} replay samples from past experiences. Thereby, samples are saved in a buffer and replayed to conduct knowledge distillation \cite{rebuffi2017icarl}. Other methods use gradients of previous tasks to mitigate gradient interference \citep{lopez2017gradient, chaudhry2018efficient}. Furthermore, generative models are used to generate synthesized samples for rehearsal \cite{shin2017continual}. \textit{Regularization} based techniques consolidate previously acquired knowledge when optimizing the model parameters. \textit{Learning without Forgetting (LwF)} \cite{li2017learning}, use knowledge distillation to regularize parameter updates. \textit{Elastic Weight Consolidation (EWC)} \cite{kirkpatrick2017overcoming} directly regularizes the parameters based on their importance concerning previous data. Similarly to EWC, in \textit{Synaptic Intelligence} \citep{zenke2017continual} the model parameter importance is regularized but in a separate parameter optimization step. \textit{Dynamic architecture} methods prevent forgetting by the deliberate reuse of \cite{mendez2021lifelong} or increase in \citep{rusu2016progressive} model parameters.



\vspace{1mm}

Concluding from the reviewed literature, this work presents, to the best of our knowledge, the first analysis on the application of continual deep-learning methods for the purpose of real-world financial statement audits.

\section{Methodology}
\label{sec:methodology}

Let \scalebox{0.9}{$X = \{x_{1}, x_{2}, x_{3}, ..., x_{N}\}$} formally define a population of \scalebox{0.9}{$i=1, 2, 3, ..., N$} journal entries. Each individual entry, denoted by \scalebox{0.9}{$x_{i} = \{x_{i}^{1}, x_{i}^{2}, ..., x_{i}^{M}; x_{i}^{1}, x_{i}^{2}, ..., x_{i}^{K}\}$}, consists of \scalebox{0.9}{$j=1, 2, 3, ..., M$} categorical accounting attributes and \scalebox{0.9}{$l=1, 2, 3, ..., K$} numerical accounting attributes. The individual attributes describe the journal entries details, e.g., the entries' posting type, posting date, amount, general-ledger.  When examining journal entries, recorded in large-scaled ERP systems, two characteristics can be observed: First, the journal entry attributes exhibit a high variety of distinct attribute values, e.g., due to the high number of vendors or distinct posting amounts. Second, journal entries exhibit a plethora of attribute value correlations, e.g., a document type usually posted in combination with a certain general ledger account. Derived from this observation we distinguish two classes of anomalous entries \cite{Breunig2000}, namely \textit{global} and \textit{local} accounting anomalies:

\vspace{1mm}

\begin{itemize}

\item \textbf{Global Anomalies} correspond to journal entries that exhibit unusual or rare individual attribute values, e.g., rarely used ledgers, or unusual posting times. Such anomalies are often simple to detect since they are created unintentionally in the context of regular posting activities. Hence, global anomalies are usually associated with a high \textit{error} risk.

\item \textbf{Local Anomalies} correspond to journal entries that exhibit unusual or rare attribute value correlations, e.g., unusual co-occurrences of general ledgers, posting types and user accounts. Such anomalies are often difficult to detect since perpetrators aim to imitate regular posting activities. Hence, local anomalies are usually associated with a high \textit{fraud} risk.

\end{itemize}

\vspace{1mm}

\noindent Ultimately, auditors aim to learn a model that detects both classes of anomalies using a continual anomaly detection framework. In the following, we describe the main components of our proposed framework. A schematic view of the framework is illustrated in Fig. \ref{fig:architecture}, encompassing an \textit{anomaly detection} and a \textit{continual} learning setup.


\vspace{1mm}

\textbf{Anomaly Detection Setup:} \citet{hawkins2002} proposed the application of \textit{Autoencoder Networks (AEN's)}, a special type of feed-forward multi-layer neural network for unsupervised anomaly detection. Formally, AEN's are comprised of two nonlinear functions referred to as \textit{encoder} and \textit{decoder} network \cite{hinton2006} that are jointly trained to reconstruct a given input. The encoder network $q_{\psi}(\cdot)$ maps an input $x_{i} \in \mathcal{R}^k$ to a code vector $z_{i} \in \mathcal{R}^m$ referred to as \textit{latent} representation, where usually $k > m$. Subsequently, this representation is mapped back by the decoder network $p_\phi(\cdot)$ to a reconstruction $\tilde{x}_{i} \in \mathcal{R}^k$ of the original input space. In an attempt to achieve $x \approx \tilde{x}$ the AEN is trained to minimize the dissimilarity of a given journal entry $x_{i}$ and its reconstruction $\tilde{x}_{i} = p_\phi(q_\psi(x_{i}))$ as faithfully as possible. Thereby, the training objective is to learn a set of optimal model parameters $\theta^{*} =\{\psi^{*} \cup \, \phi^{*}\}$ by minimizing the AEN's reconstruction loss ($\mathcal{L}_{Rec}$), formally denoted as:

\begin{equation}
    \arg \min_{\phi, \psi} \|x_{i} - p_{\phi}(q_{\psi}(x_{i}))\|,
    \label{equ:reconstruction_loss}
\end{equation}

\noindent where $i=1, 2, 3,..., N$. Upon successful AEN training, the model is used to obtain the individual reconstruction error of each journal entry. In a real-world audit scenario, entries that correspond to a high reconstruction error will then be selected for subsequent substantial examination.

\vspace{1mm}


\textbf{Continual Learning Setup:} In \textit{Continual Learning (CL)} a model $f_{\theta}$, e.g., an autoencoder $f_{\theta}\!: q_{\psi} \oplus p_{\phi}$, is trained in a setup where data \scalebox{0.8}{$D=\bigcup_{i=1}^{M} D_{i}$} is observed as a stream of $M$ disjoint experiences \scalebox{0.9}{$\{E_{i}\}_{i=1}^{M}$}. In the context of auditing, the journal entry data $D_{i}$ of a particular experience $E_i$ originates from $N$ organizational posting activities $A_{j}$ as defined by \scalebox{0.8}{$D_i=\bigcup_{j=1}^{N} A_{j}$}. A model trained on the data of the \textit{i}-th experience is denoted as \scalebox{0.85}{$f^{(i)}_{\theta}$}, where $\theta$ defines the model parameters. The models reconstruction loss can then be derived by:

\begin{equation}
    \mathcal{L}_{Rec}^{(i)} = \sum_{j=1}^{N} {\mathcal{L}_{Rec}(f_{\theta}^{(i)}, A_j)},
    \label{equ:per_dept_loss}
\end{equation}

\noindent where $A_{j}$ denotes the organizational posting activities of experience $E_{i}$.

\vspace{1mm}

In this work three CL setups are utilized to audit financial accounting data, namely (i) \textit{Sequential Fine-tuning}, (ii) \textit{EWC}, and (iii) \textit{Experience Replay}. In each setup an anomaly detection model is incrementally trained on a stream of $M-1$ past experiences $\{E_{i}\}_{i=1}^{M-1}$. For each experience $E_{i}$, the model parameters $\theta$ are initialized with the optimal model parameter learned at the previous experience $E_{i-1}$. In the \textit{Sequential Fine-Tuning (SFT)} setup, the parameters $\theta$ are optimized, as defined by:
\vspace{-1mm}

\begin{equation}
     \theta^{* (i)} = \arg\min_{\theta}{\mathcal{L}(f(\theta), D_i| \theta_{init}=\theta^{*(i-1)})},
     \label{equ:sequential_fine_tuning}
\end{equation}

\noindent where $\theta^{* (i)}$ denotes the optimal parameters obtained for the previous experience $E_i$. In the regularization based \textit{Elastic Weight Consolidation (EWC)} setup, the parameters $\theta$ are optimized, as defined by:
\vspace{-1mm}

\begin{equation}
    \theta^{* (i)}= \arg\min_{\theta}{\widehat{\mathcal{L}}(f(\theta), D_i| \theta_{init}=\theta^{*(i-1)})},
    \label{equ:elastic_weight_consolidation}
\end{equation}

\noindent where $\theta^{* (i)}$ denotes the optimal parameters obtained for the previous experience $E_i$. To update the model's parameters in a given experience, EWC \cite{kirkpatrick2017overcoming} is applied by determining the Fisher information matrix $F$ according to \scalebox{0.9}{$\hat{\mathcal{L}}(\theta) =  \mathcal{L}_{Rec}(\theta) + \sum_{k}{\frac{\lambda}{2}}F_k(\theta_i - \tilde{\theta}_{k})^2$}, where $k$ denotes the model parameters of the current model and \scalebox{0.9}{$\tilde{\theta}$} the optimal parameters of the previous experience. In the rehearsal based \textit{Experience Replay (ER)} setup, the parameters $\theta$ are optimized, as defined by:
\vspace{-1mm}

\begin{equation}
     \theta^{* (i)} = \arg\min_{\theta}{\mathcal{L}(f(\theta), \widehat{D}_i| \theta_{init}=\theta^{*(i-1)})},
    \label{equ:experience_replay}
\end{equation}

\noindent where $\theta^{* (i)}$ denotes the optimal parameters obtained for the previous experience $E_i$. To update the model's parameters in a given experience, samples of past experiences are replayed. Therefore, a sample buffer $\mathcal{B}$ of capacity $N_B$ is employed. The buffer comprises an equal number of \scalebox{0.9}{$\lfloor \frac{N_B}{M} \rfloor$} samples for each observed experience. 

\vspace{1mm}

We evaluate the continual learning setups against two baselines, namely (i) \textit{single experience} and (ii) \textit{joint experience} learning. In \textit{Single Experience Learning (SEL)} only the journal entries of the current in-scope audit experience $E_M$ are used to train $f_{\theta}$. Such a learning setup is commonly applied in contemporary audits. The model parameters $\theta$ are randomly initialized and optimized, as defined by:
\vspace{-1mm}

\begin{equation}
    \theta^{* (i)} = \arg\min_{\theta}{\mathcal{L}(f(\theta), D_i| \theta_{init}={Rand})},
    \label{equ:obj_joint_training}
\end{equation}

\noindent where $\theta^{* (i)}$ denotes the optimal parameters obtained after training on $E_i$. We view this setup as performance upper-bound in our experiments. In \textit{Joint Experience Learning (JEL)} the journal entries of all previous experiences \scalebox{0.9}{$\{E_{i}\}_{i=1}^{M-1}$} and the current in-scope experience \scalebox{0.9}{$E_{M}$} are used to train $f_{\theta}$ from scratch. However, such a learning setup is not desirable in contemporary audit engagements due to time-budget constraints. The model parameters $\theta$ are randomly initialized and optimized, as defined by:
\vspace{-1mm}

\begin{equation}
    \theta^{* (i)} = \arg\min_{\theta}{\mathcal{L}(f(\theta), \{D_1, ..., D_i\}| \theta_{init}={Rand})},
    \label{equ:obj_joint_training}
\end{equation}

\noindent where $\theta^{* (i)}$ denotes the optimal parameters obtained after training on $E_i$. Nowadays, this setup is commonly rejected due to time-budget constraints of audit engagements. We view this setup as performance lower-bound in our experiments. 



\vspace{-2mm}

\section{Experimental Setup}
\label{sec:experiments}

In this section, we describe the dataset, data pre-processing, audit scenarios and experimental details to continually learn from real-world accounting data.

\begin{figure*}[ht!]
    \begin{subfigure}{0.247\textwidth}
        \centering
        \includegraphics[height=2.8cm, trim=0 0 0 0, clip]{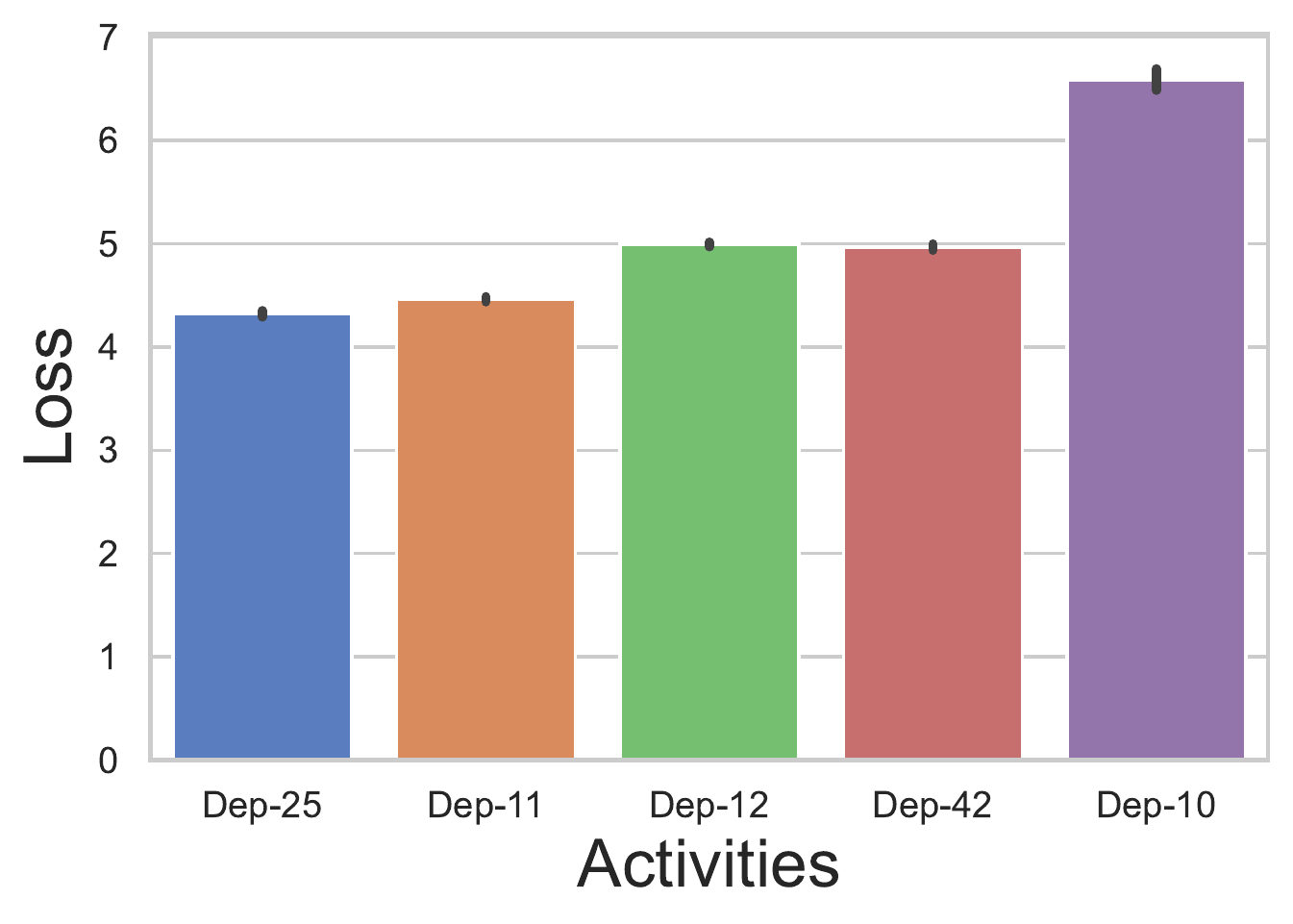}
        \vspace{-2mm}
        \caption{\scalebox{0.82}{Single Experience Learning (SEL)}}
    \end{subfigure}
    \begin{subfigure}{0.247\textwidth}
        \centering
        \includegraphics[height=2.8cm, trim=0 0 0 0, clip]{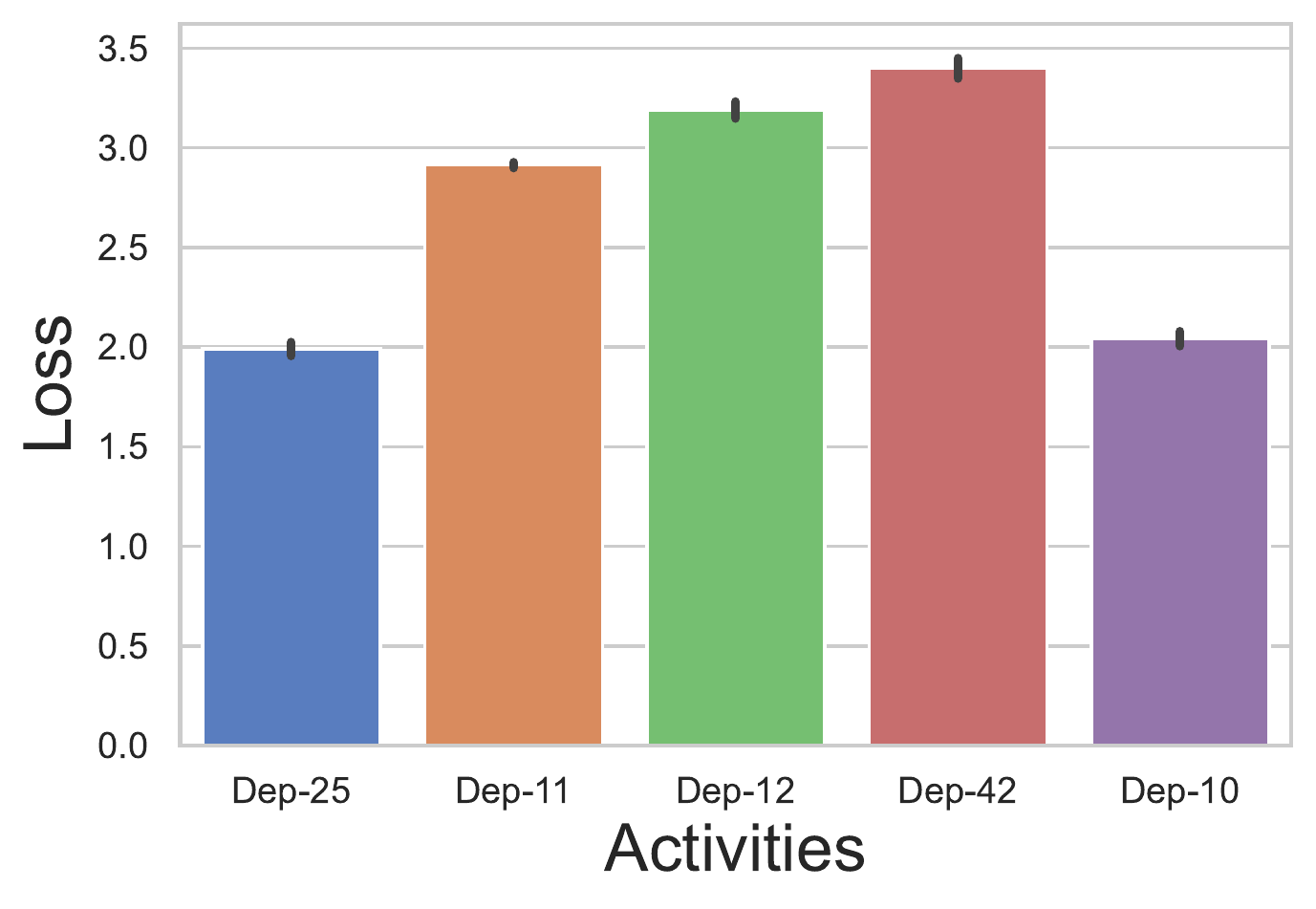}
        \vspace{-2mm}
        \caption{\scalebox{0.82}{Joint Experience Learning (JEL)}}
    \end{subfigure}
    \begin{subfigure}{0.247\textwidth}
        \centering
        \includegraphics[height=2.8cm, trim=0 0 0 0, clip]{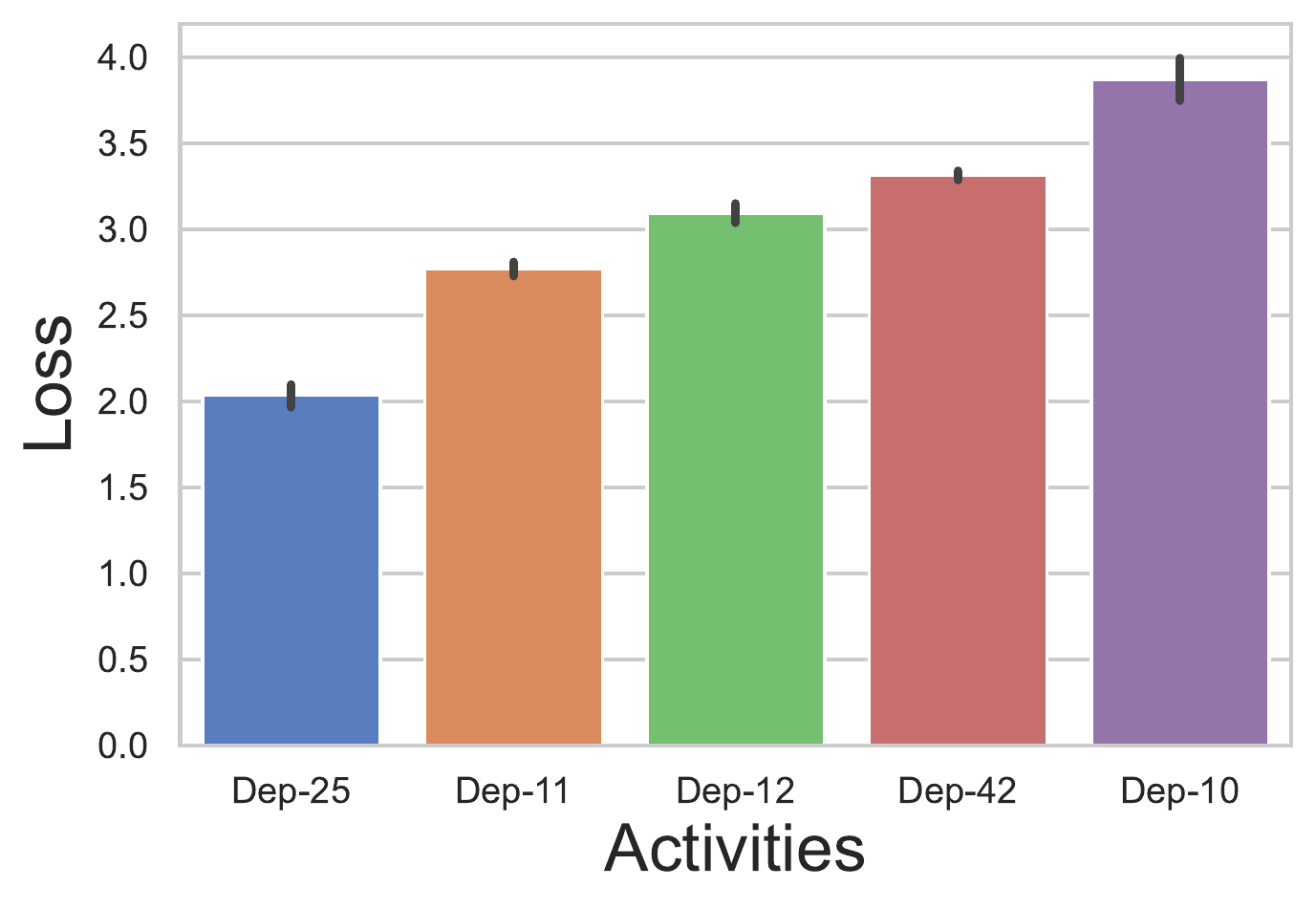}
        \vspace{-2mm}
        \caption{\scalebox{0.82}{Sequential Fine-Tuning (SFT)}}
    \end{subfigure}
    \begin{subfigure}{0.247\textwidth}
        \centering
        \includegraphics[height=2.8cm, trim=0 0 0 0, clip]{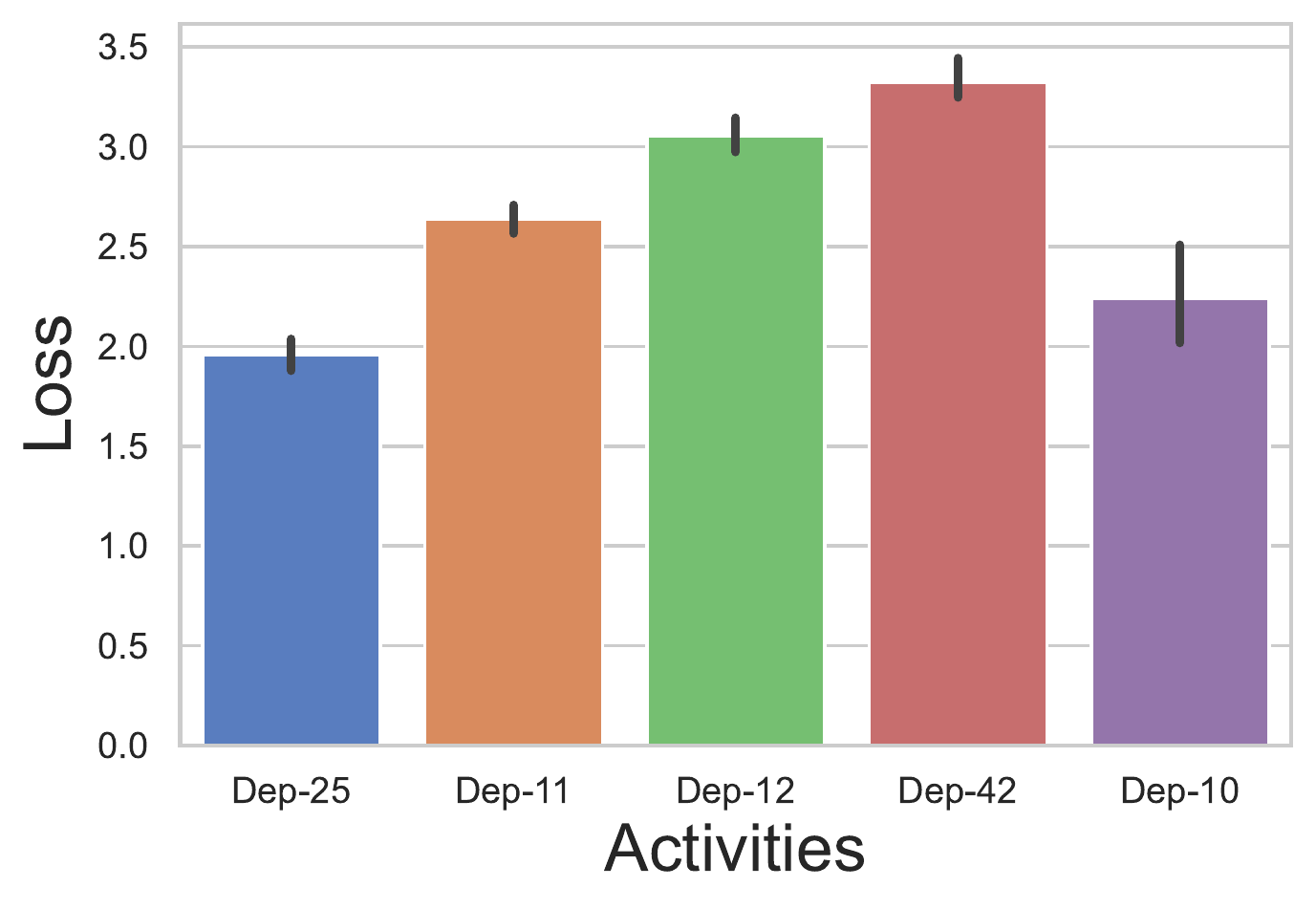}
        \vspace{-2mm}
        \caption{\scalebox{0.82}{Experience Replay (ER)}}
    \end{subfigure}
    \vspace{-2mm}
    \caption{Average reconstruction losses $\mathcal{L}_{Rec}$ in the in-scope audit period $E_{M}$ of dataset $D^{A}$ for the decaying target department (purple bar) and the four departments exhibiting the highest loss (non-purple bars). It can be observed that in CL the linearly decaying target department \scalebox{0.95}{$A'$} results in a significantly reduced number of loss magnitude based false-positive alerts.}
    \label{fig:barplots}
    \vspace{-5mm}
\end{figure*}

\vspace{1mm}

\textbf{Datasets and Data Preparation: } To evaluate the learning capabilities of the proposed continual learning framework we use two publicly available datasets of real-world financial payment data.\footnote{Due to the general confidentiality of journal entry data, we evaluate the proposed methodology based on two public available real-world datasets to allow for the reproducibility of our results.} The datasets exhibit high similarity to real-world ERP accounting data, e.g., typical manual payments or payment run records. 

\begin{itemize}
\item The \textit{City of Philadelphia} payments\footnote{\url{https://www.phila.gov/2019-03-29-philadelphias-initial-release-of-city-payments-data/}} denoted as \scalebox{0.9}{$D^{A}$} encompass a total of 238,894 payments generated by 58 distinct city departments and comprises $10$ categorical and $1$ numerical attribute(s).

\item The \textit{City of Chicago} payments\footnote{\url{https://data.cityofchicago.org/Administration-Finance/Payments/s4vu-giwb/}} denoted as \scalebox{0.9}{$D^{B}$} encompass a total of 399,158 payments generated by 54 distinct city departments and comprises of $6$ categorical and $1$ numerical attribute(s).
\end{itemize}

For both datasets, let \scalebox{0.9}{$\{A_{j}\}_{j=1}^{N}$} denote the set of payments generating city departments. For each dataset we conduct the following pre-processing steps: First, the $\tau$ departments exhibiting the highest number of payments \scalebox{0.9}{$\{A^{*}_{k}\}_{k=1}^{\tau}$} are determined. Second, for each city department \scalebox{0.9}{$A^{*}_{k}$} a set of \scalebox{0.9}{$\eta$} payments is randomly sampled \scalebox{0.9}{$x_{i} \sim \{A^{*}\}_{k=1}^{\tau}$}, where \scalebox{0.9}{$i=1,..,\eta$}. Third, the sampled payment's attribute values are transformed according to:

\begin{itemize}

    \item The categorical attribute values of \scalebox{0.9}{$x_{i}^{j}$} are converted into \textit{one-hot} numerical tuples of bits \scalebox{0.9}{$\hat{x}_{i}^{j} \in \{0, 1\}^{\upsilon}$}, where $\upsilon$ denotes the number of unique attribute values in \scalebox{0.9}{$x^{j}$}.
    
    \item The numerical attribute values of \scalebox{0.9}{$x_{i}^{l}$} are scaled, according to \scalebox{0.9}{$\hat{x}_{i}^{l} = (x_{i}^{l} - \text{min}(x^{l}) / (\text{max}(x^{l}) - \text{min}(x^{l}))$}, where the $\text{min}$ and $\text{max}$ are obtained over all attribute values in \scalebox{0.9}{$x^{l}$}. 
    
\end{itemize}

 \noindent We set the number of departments \scalebox{0.9}{$\tau = 10$} and number of samples \scalebox{0.9}{$\eta=10,000$} in all our experiments to derive the pre-processed journal entries \scalebox{0.9}{$\hat{x}\in \hat{X}$}, where \scalebox{0.9}{$\hat{x}_{i} = \{\hat{x}_{i}^{1}, \hat{x}_{i}^{2}, ..., \hat{x}_{i}^{M}; \hat{x}_{i}^{1}, \hat{x}_{i}^{2}, ..., \hat{x}_{i}^{K}\}$}. Ultimately, each payment \scalebox{0.9}{$\hat{x}_{i} \in D^{A}$} resulted in \scalebox{0.9}{5,588} encoded dimensions, while each payment \scalebox{0.9}{$\hat{x}_{i} \in D^{B}$} in \scalebox{0.9}{7,486} encoded dimensions.

\vspace{1mm}

\begin{figure}[t!]
	\hspace*{0.0cm} \includegraphics[width=8.2cm, angle=0, trim={0.0 0.0 0.0 0.0}]{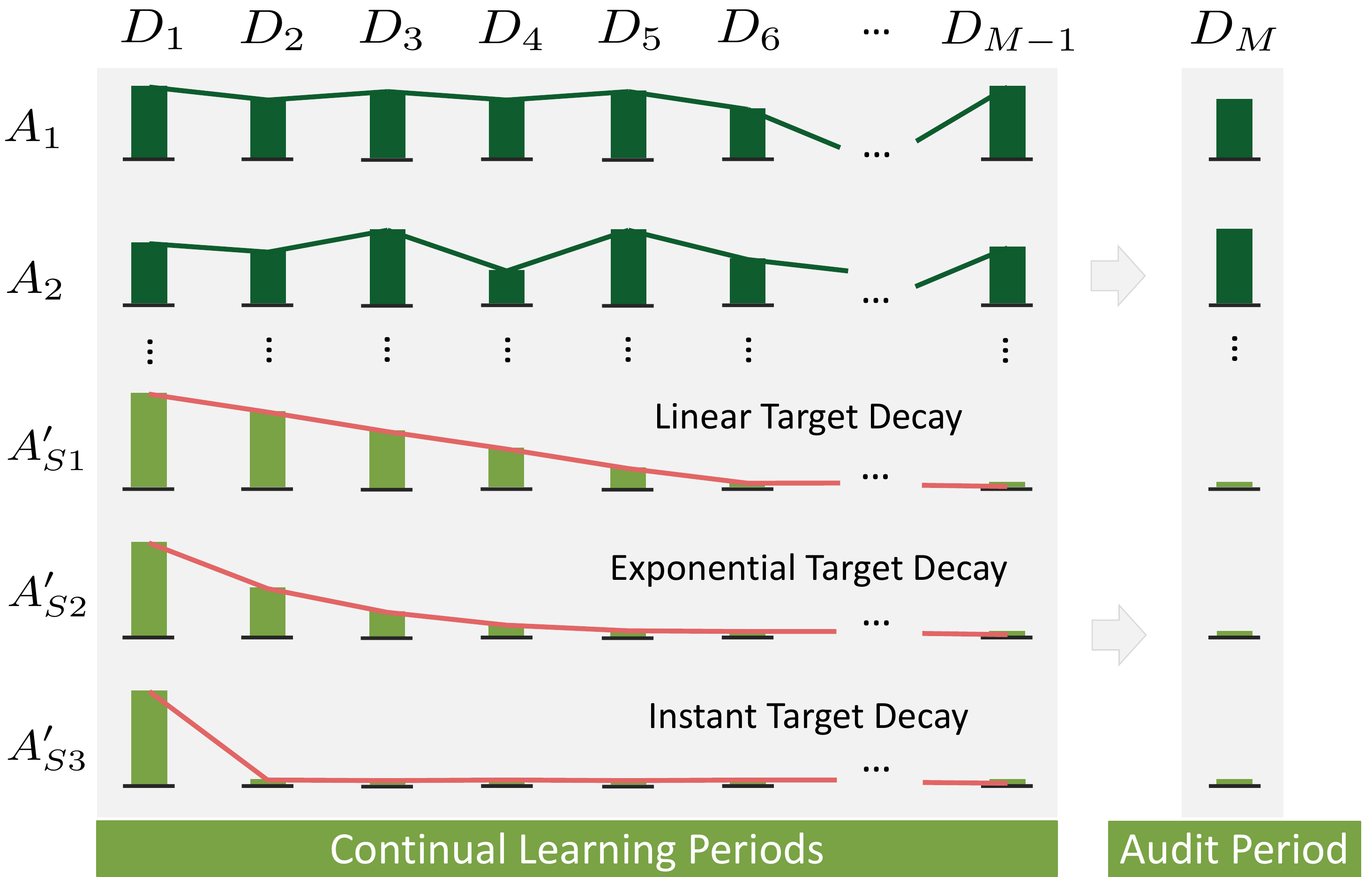}
	\vspace{-1mm}
	\caption{Journal entry distribution for distinct activities \scalebox{0.9}{$A_{1}, A_{2}, ..., A_{\tau}$} over progressing experiences (top) and target department decaying scenarios \scalebox{0.9}{$A_{S1}', A_{S2}', A_{S3}'$} (bottom).} 
	\label{fig:dataset}
	\vspace{-5mm}
\end{figure}

\textbf{Continual Audit Scenarios: } We create a stream of $N$ experiences denoted by \scalebox{0.9}{$\{E\}_{i=1}^{M}$} where each experience exhibits its own dataset \scalebox{0.9}{$D_{i}$}. For each $D_{i}$ a number of \scalebox{0.9}{$\rho \cdot \frac{\eta}{N}$} pre-processed payments is randomly sampled from each $\hat{x}_{i} \sim A^{*}_{k}$, where $\rho \in [0.9;1.0]$ and $k=1,..,\tau$. In a real-world audit setup the experiences may correspond to distinct time periods, e.g., a financial quarter or year. Furthermore, we assume that the most recent experience $E_{M}$ defines the in-scope period of the audit with journal entry data $D_{N}$. In this initial work, we evaluate two scenarios observable when examining journal entry data in financial audits, namely (i) \textit{discontinued operations} and (ii) \textit{anomalous postings}. Figure \ref{fig:dataset} illustrates the distribution of sampled journal entries for distinct activities over progressing experiences. 

\vspace{1mm}

\begin{figure*}[ht!]
    \begin{subfigure}{0.33\textwidth}
        \centering
        \includegraphics[height=3.0cm, trim=0 0 0 0, clip]{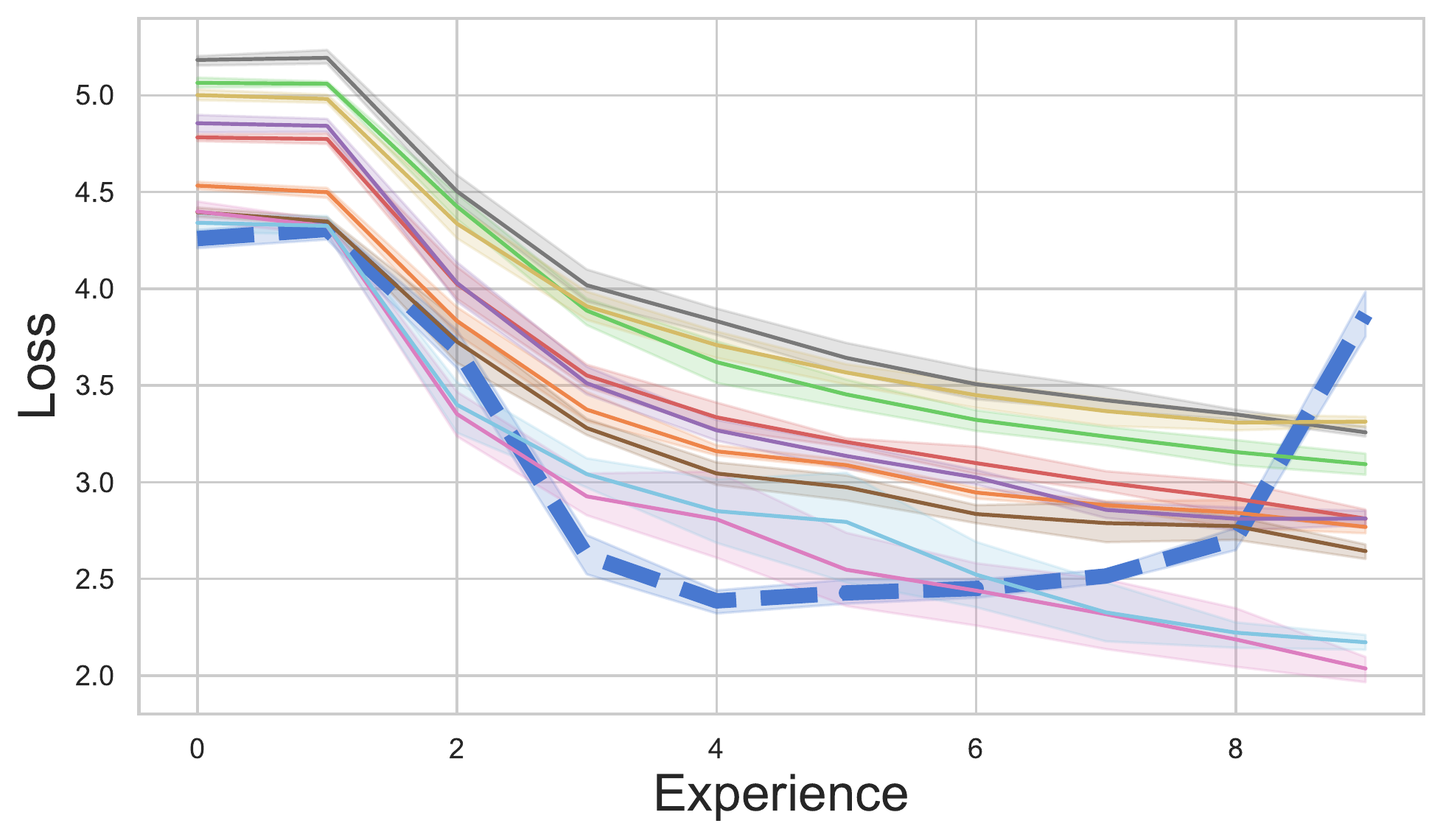}
        \vspace{-2mm}
        \caption{\scalebox{0.82}{Sequential Fine-Tuning (SFT)}}
    \end{subfigure} \hspace{-0.3cm}
    \begin{subfigure}{0.33\textwidth}
        \centering
        \includegraphics[height=3.0cm, trim=0 0 0 0, clip]{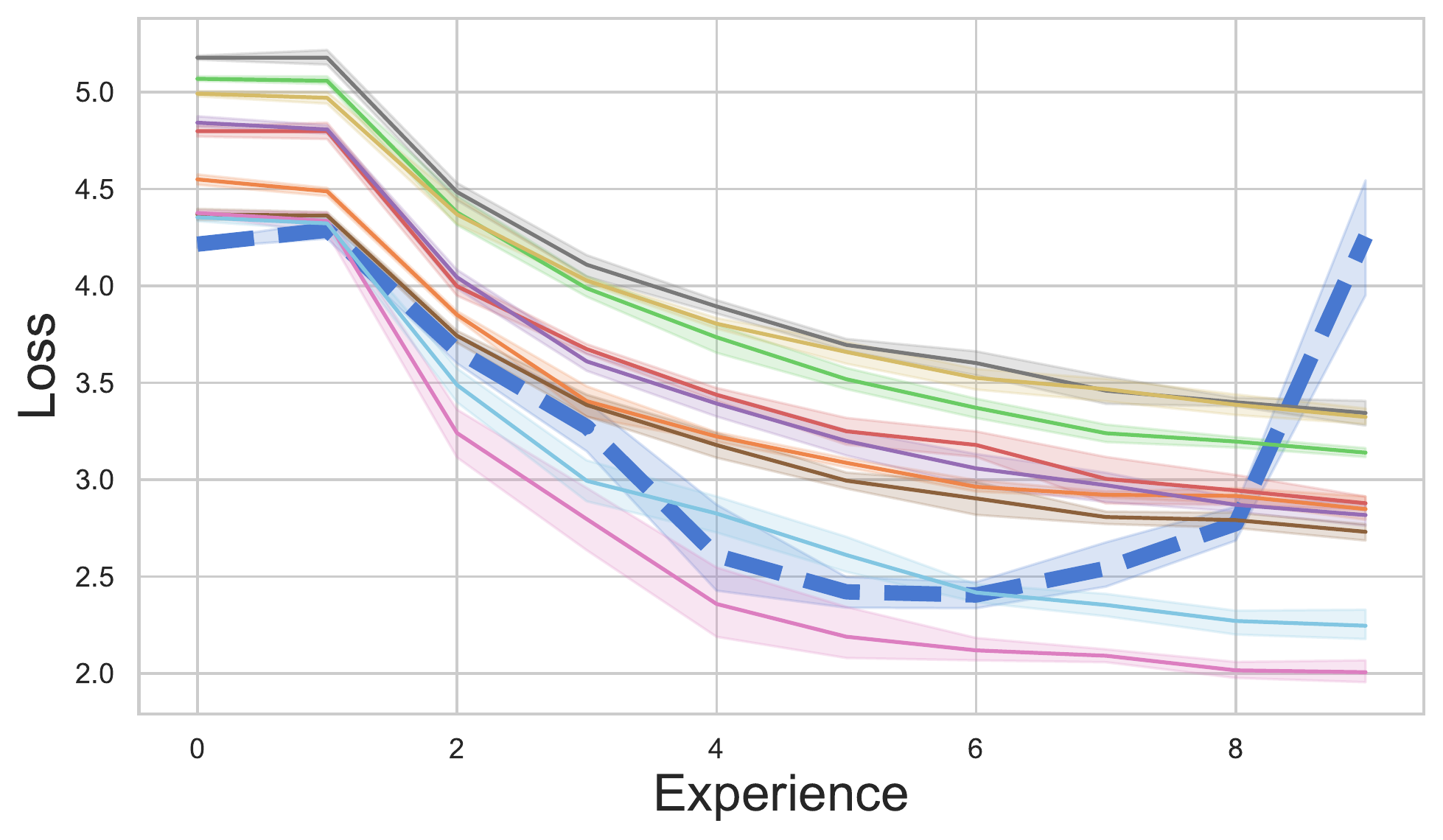}
        \vspace{-2mm}
        \caption{\scalebox{0.82}{Elastic Weight Consolidation (EWC)}}
    \end{subfigure}
    \begin{subfigure}{0.33\textwidth}
        \centering
        \includegraphics[height=3.0cm, trim=0 0 0 0, clip]{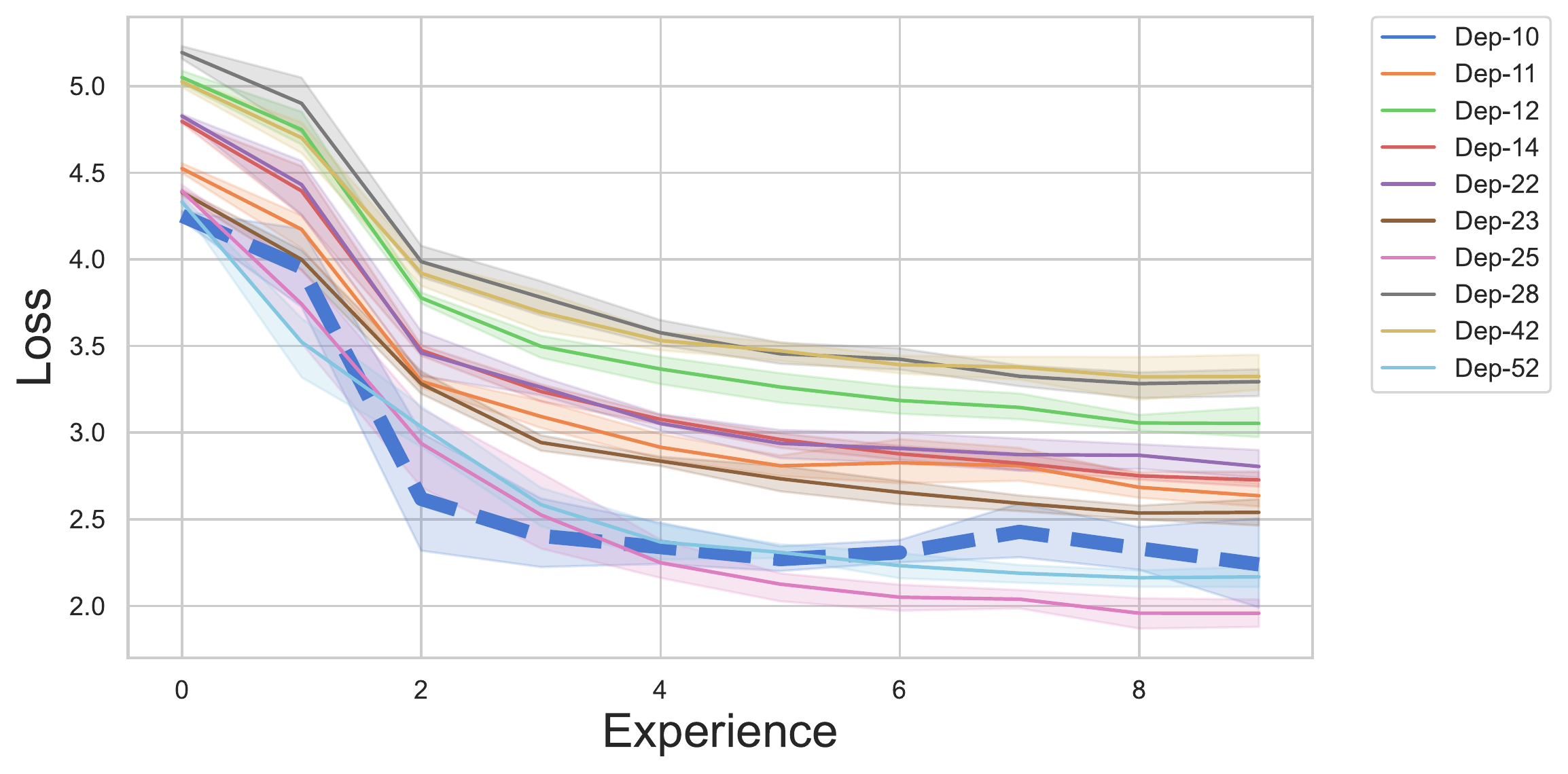}
        \vspace{-6mm}
        \caption{\scalebox{0.82}{Experience Replay (ER)}}
    \end{subfigure}
    \vspace{-2mm}
    \caption{Reconstruction losses $\mathcal{L}_{Rec}$ of the $\tau=10$ departments in dataset $D_A$ over a stream of \scalebox{0.9}{$\{E\}_{i=1}^{M}$} experiences for distinct CL setups. The dashed blue line corresponds to reconstruction loss of the linearly decaying target department \scalebox{0.95}{$A'$}. It can be observed that ER results in a significantly reduced number of false-positives over the progressing experiences. }
    \label{fig:lineplots}
    \vspace{-5mm}
\end{figure*}

The first \textit{discontinued operations} scenario, refers to a situation in which an organizational activity or department is discontinued, e.g., due to a shut down a divestiture. Thereby, the number of corresponding journal entries vanishes over time. Using stationary deep-learning enabled audit techniques such a scenario regularly results in a high number of \textbf{false-positive alerts} within the in-scope audit period. To establish such a scenario a to be discontinued department \scalebox{0.9}{$A' \in \{A^{*}_{k}\}_{k=1}^{\tau}$} is selected. Subsequently, a decaying number of journal entries is randomly sampled from \scalebox{0.9}{$A'$} with progressing experiences. We investigated three discontinuation scenarios, resulting in a (i) \textit{linear} (\scalebox{0.9}{$A_{S1}'$}), (ii) \textit{exponential} (\scalebox{0.9}{$A_{S2}'$}), and (iii) \textit{instant} (\scalebox{0.9}{$A_{S3}'$}) reduction of journal entries.


\vspace{1mm}

The second \textit{anomalous postings} scenario, refers to a situation in which \textit{`anomalous'} journal entries are posted within the in-scope period of an audit, e.g., due to error or fraudulent activities. Using stationary deep-learning enabled audit techniques such a scenario regularly results in an increased number of undetected \textbf{false-negative observations} within the in-scope audit period. To establish such a scenario, a small number of $\alpha_{1}$ local (\scalebox{0.9}{$A_{P1}$}) and $\alpha_{2}$ global (\scalebox{0.9}{$A_{P2}$}) accounting anomalies is injected into the dataset $D_{N}$ of the last experience.\footnote{To create both classes of anomalies, we built upon the \textit{Faker} project, which is publicly available via the following URL: \url{https://github.com/joke2k/faker}.} For both classes of of injected anomalies we set $\alpha_{1}=10$ and $\alpha_{2}=10$ in all our experiments.



\vspace{1mm}

\textbf{Autoencoder Learning Setup:} The encoder $q_{\psi}$ and a decoder $p_{\theta}$ network use Leaky-ReLU activation functions \cite{xu2015} with scaling factor $\alpha = 0.4$ except in the bottleneck (last) layer where tanh linearities (non-linearities) are applied. Table \ref{tab:architecture} depicts the architectural details of the models used in our experiments. Both AEN parameter sets $\psi$ and $\phi$ are initialized as described by \citet{glorot2010}. The models are trained for 500 epochs, using a batch size of $b=128$ journal entries, and early stopping once the loss converges. We use Adam optimization \cite{kingma2014} with $\beta_{1}=0.9$, $\beta_{2}=0.999$. Furthermore, we compute the binary cross-entropy error of a given encoded journal entry $\hat{x}_{i}$, as defined by \scalebox{0.9}{$\mathcal{L}_{Rec}(\tilde{x}_{i}, \hat{x}_{i}) = \frac{1}{N} \sum_{i=1}^{N} \tilde{x}_{i} \cdot \log(\hat{x}_{i}) + (1 - \tilde{x}_{i}) \cdot \log(1 - \hat{x}_{i})$}, where \scalebox{0.9}{$\tilde{x}_{i} = p_{\phi}(q_{\psi}(\hat{x}_{i}))$} denotes the i-\textit{th} reconstructed entry.

\vspace{-2mm}

 \begin{table}[ht!]
  \caption{Number of neurons per layer $\eta$ of the encoder $q_{\psi}$ and decoder $p_{\phi}$ network that constitute the distinct AEN architectures used in our experiments.} 
    \vspace*{-2mm}
  \fontsize{8}{6}\selectfont
  \centering
  \begin{tabular}{l c c c c c c c}
    \toprule
        \multicolumn{1}{l}{Net}
        & \multicolumn{1}{c}{$\eta$ = 1}
        & \multicolumn{1}{c}{2}
        & \multicolumn{1}{c}{3}
        & \multicolumn{1}{c}{4}
        & \multicolumn{1}{c}{5}
        & \multicolumn{1}{c}{6}
        & \multicolumn{1}{c}{7}
        \\
    \midrule
    $q_{\psi}(z|\hat{x})$ & 128 & 64 & 32 & 16 & 8 & 4 & 2 \\
    $p_{\phi}(\hat{x}'|z)$ & 2 & 4 & 8 & 16 & 32 & 64 & 128 \\
    \bottomrule \\
  \end{tabular}
    \label{tab:architecture}
    \vspace*{-4mm}
 \end{table} 

\textbf{Continual Learning Setup:} We set the number of experiences \scalebox{0.9}{$N=10$} in all our experiments. To find optimal parameters in EWC and ER we conducted a grid search of the hyper-parameter space. For EWC we found that $\lambda=50$ preserves an optimal degree of plasticity. In ER learning we set $N_{B}=500$ which seemed sufficient for both datasets. We reset the optimizer parameters upon each experience to avoid past experience information transfer through the optimizer state. We used Avalanche \cite{lomonaco2021avalanche} to implement our experiments and published the code on GitHub. \footnote{ \url{https://github.com/GitiHubi/deepContinualAuditing}}

\vspace{-2mm}


\section{Experimental Results and Discussion}
\label{sec:results}
In this section, we present the experimental continual learning results for both continual audit scenarios, namely (i) \textit{discontinued operations} and (ii) \textit{anomalous postings}.

\vspace{1mm}

The experimental results of the \textit{discontinued operations} scenario are presented in Tab. \ref{tab:reconstruction_scores_philadelphia}. For both datasets $D^{A}$ and $D^{B}$ it can be observed that the average reconstruction loss difference of the target department $A'$ and the department exhibiting the highest loss $\tilde{A}$, as defined by \scalebox{0.9}{$\Delta_{FP} := \mathcal{L}_{Rec}(A_{S}') - \mathcal{L}_{Rec}(\tilde{A})$}. A negative (positive) $\Delta_{FP}$ indicates a reduction (an increase) of false-positive alerts originating from the decaying target department within the in-scope audit period $E_{M}$. Thereby, JEL and SEL provide a lower- and upper-bound on the false-positive alerts. The obtained results demonstrate that CL mitigates the risk of false-positive alerts while saving data and computational resources. When comparing SFT, EWC, and ER for the decaying setups, as also illustrated in Fig. \ref{fig:lineplots}, the best performance is achieved by ER. Thereby, the exponential $A_{S2}'$ and instant $A_{S3}'$ decay results in a more challenging CL setup. 

\vspace{-2mm}

\begin{table}[ht!]
\caption{Average reconstruction loss $\mathcal{L}_{Rec}$ difference $\Delta_{FP}$ at $E_{M}$ between the decaying target department and the department with the highest loss in both datasets $D^{A}$ and $D^{B}$.} 
\fontsize{9}{7}\selectfont
\centering
\vspace{-2mm}
\begin{tabular}{l r r r}
    \multicolumn{4}{l}{Discontinued Operations - Philadelphia Payments $D^{A}$} \\ 
    \vspace{-1mm} \\
    \multicolumn{1}{l}{Method}
    & \multicolumn{1}{c}{$\Delta_{FP}(A_{S1}')$}
    & \multicolumn{1}{c}{$\Delta_{FP}(A_{S2}')$}
    & \multicolumn{1}{c}{$\Delta_{FP}(A_{S3}')$}
    \\
\toprule
\scalebox{0.9}{SEL} & 1.45 $\pm$ 0.115 & 1.40 $\pm$ 0.064 & 1.45 $\pm$ 0.062 \\
\scalebox{0.9}{JEL} & -1.36 $\pm$ 0.047 & -1.36 $\pm$ 0.047 & -1.36 $\pm$ 0.047 \\
\midrule
\scalebox{0.9}{SFT} & 0.56 $\pm$ 0.112 & 1.39 $\pm$ 0.401 & 1.82 $\pm$ 0.284 \\
\scalebox{0.9}{EWC} & 0.91 $\pm$ 0.400 & 1.31 $\pm$ 0.313 & 2.03 $\pm$ 0.218 \\
\scalebox{0.9}{ER} & \textbf{-1.08} $\pm$ \textbf{0.371} & \textbf{0.02} $\pm$ \textbf{0.539} & \textbf{0.02} $\pm$ \textbf{0.089} \\
\bottomrule \\
\multicolumn{4}{l}{\scalebox{0.8}{Stdv. $\pm$ originates from parameter initialization using five random seeds.}}
\end{tabular}
\label{tab:reconstruction_scores_philadelphia}
\vspace*{-2mm}
\end{table}

\begin{table}[ht!]
\fontsize{9}{7}\selectfont
\centering
\begin{tabular}{l r r r}
    \multicolumn{4}{l}{Discontinued Operations - Chicago Payments $D^{B}$} \\ 
    \vspace{-1mm} \\
    \multicolumn{1}{l}{Method}
    & \multicolumn{1}{c}{$\Delta_{FP}(A_{S1}')$}
    & \multicolumn{1}{c}{$\Delta_{FP}(A_{S2}')$}
    & \multicolumn{1}{c}{$\Delta_{FP}(A_{S3}')$}
    \\
\toprule
\scalebox{0.9}{SEL} & 0.66 $\pm$ 0.101 & 0.66 $\pm$ 0.053 & 0.61 $\pm$ 0.099 \\
\scalebox{0.9}{JEL} & -0.96 $\pm$ 0.153 & -0.96 $\pm$ 0.153 & -0.96 $\pm$ 0.153 \\
\midrule
\scalebox{0.9}{SFT} & 0.78 $\pm$ 0.268 & 1.71 $\pm$ 1.207 & 1.35 $\pm$ 0.213 \\
\scalebox{0.9}{EWC} & 1.18 $\pm$ 0.168 & 1.12 $\pm$ 0.184 & 1.10 $\pm$ 0.230 \\
\scalebox{0.9}{ER} & \textbf{-0.05} $\pm$ \textbf{1.514} & \textbf{-0.50} $\pm$ \textbf{0.192} & \textbf{-0.19} $\pm$ \textbf{0.243} \\
\bottomrule \\
\multicolumn{4}{l}{\scalebox{0.8}{Stdv. $\pm$ originates from parameter initialization using five random seeds.}}
\end{tabular}
\label{tab:reconstruction_scores_chicago}
\vspace*{-5mm}
\end{table}

The experimental results of the \textit{anomalous postings} scenario and a linear decaying target department are presented in Tab. \ref{tab:anomaly_scores_philadelphia}. For both datasets $D^{A}$ and $D^{B}$ it can be observed that SEL, the average reconstruction error $\mathcal{L}_{Rec}$ of the decaying target department $A'$ and the local anomalies $A_{P1}$ exhibit a similar error magnitude within the in-scope audit period $E_{M}$. Such a scenario increases the risk of not detecting local anomalies and might result in false-negative decisions. The obtained results demonstrate that CL mitigates this risk by forward transfer of knowledge from previous experiences. This observation is also reflected by the loss difference \scalebox{0.9}{$\Delta_{FN} := \mathcal{L}_{Rec}(A') - \mathcal{L}_{Rec}(A_{P1})$} of the target department $A'$ and the local anomalies $A_{P1}$. When comparing SFT, EWC, and ER, ER achieves the best performance.

\vspace{-2mm}

\begin{table}[ht!]

\caption{Average reconstruction loss $\mathcal{L}_{Rec}$ at $E_{M}$ of the decaying target department $A'$ and the injected local $A_{P1}$ and global $A_{P2}$ anomalies in both datasets $D^{A}$ and $D^{B}$.} 
\fontsize{9}{7}\selectfont
\centering
\vspace{-2mm}
\begin{tabular}{l r r r}
    \multicolumn{4}{l}{Anomalous Postings - Philadelphia Payments $D^{A}$} \\ 
    \vspace{-1mm} \\
    \multicolumn{1}{l}{Method}
    & \multicolumn{1}{c}{$\mathcal{L}_{Rec} (A') \downarrow$}
    & \multicolumn{1}{c}{$\mathcal{L}_{Rec} (A_{P1}) \uparrow$}
    & \multicolumn{1}{c}{$\mathcal{L}_{Rec} (A_{P2}) \uparrow$}
    \\
\toprule
\scalebox{0.9}{SEL} & 6.59 $\pm$ 0.125 & 7.03 $\pm$ 0.130 & 12.21 $\pm$ 0.208 \\
\scalebox{0.9}{JEL} & 2.04 $\pm$ 0.042 & 36.71 $\pm$ 5.948 & 18.12 $\pm$ 7.399 \\
\midrule
\scalebox{0.9}{SFT} & 3.87 $\pm$ 0.141 & 34.10 $\pm$ 6.025 & 12.99 $\pm$ 2.286 \\
\scalebox{0.9}{EWC} & 4.25 $\pm$ 0.339 & 29.97 $\pm$ 4.761 & 13.22 $\pm$ 1.145 \\
\scalebox{0.9}{ER} & \textbf{2.24} $\pm$ \textbf{0.292} & \textbf{47.69} $\pm$ \textbf{6.482} & \textbf{16.64} $\pm$ \textbf{5.176} \\
\bottomrule \\
\multicolumn{4}{l}{\scalebox{0.8}{Stdv. $\pm$ originates from parameter initialization using five random seeds.}}
\end{tabular}
\label{tab:anomaly_scores_philadelphia}
\vspace*{-5mm}
\end{table}

\begin{table}[ht!]
\fontsize{9}{7}\selectfont
\centering

\vspace{-3mm}

\begin{tabular}{l r r r}
    \multicolumn{4}{l}{Anomalous Postings - Chicago Payments $D^{B}$} \\ 
    \vspace{-1mm} \\
    \multicolumn{1}{l}{Method}
    & \multicolumn{1}{c}{$\mathcal{L}_{Rec} (A') \downarrow$}
    & \multicolumn{1}{c}{$\mathcal{L}_{Rec} (A_{P1}) \uparrow$}
    & \multicolumn{1}{c}{$\mathcal{L}_{Rec} (A_{P2}) \uparrow$}
    \\
\toprule
\scalebox{0.9}{SEL} & 4.27 $\pm$ 0.097 & 3.55 $\pm$ 0.171 & 4.53 $\pm$ 0.053 \\
\scalebox{0.9}{JEL} & 1.30 $\pm$ 0.145 & 16.75 $\pm$ 4.036 & 4.71 $\pm$ 0.454 \\
\midrule
\scalebox{0.9}{SFT} & 3.01 $\pm$ 0.288 & 13.03 $\pm$ 3.845 & 3.93 $\pm$ 0.169 \\
\scalebox{0.9}{EWC} & 3.42 $\pm$ 0.185 & 12.20 $\pm$ 1.958 & 5.18 $\pm$ 1.709 \\
\scalebox{0.9}{ER} & \textbf{2.26} $\pm$ \textbf{1.154} & \textbf{22.23} $\pm$ \textbf{5.439} & \textbf{6.25} $\pm$ \textbf{1.857} \\
\bottomrule \\
\multicolumn{4}{l}{\scalebox{0.8}{Stdv. $\pm$ originates from parameter initialization using five random seeds.}}
\end{tabular}
\vspace*{-6mm}
\end{table}


\vspace{-1mm}
\section{Conclusion}
\label{sec:conclusion}

In this work, we proposed the first CL inspired framework to detect accounting anomalies in the context of financial statement audits. Based on two deliberately designed audit scenarios, we demonstrated the ability of our method to continuously learn data distribution shifts observable in datasets of real-world city payments. Our experimental results provide initial evidence that CL enabled anomaly detection improve stationary baseline detection techniques in terms of false-positive alerts and false-negative decisions. 

\newpage

\bibliography{library}

\end{document}